\pdfoutput=1

\documentclass[11pt]{article}

\usepackage{acl} 

\usepackage{times}
\usepackage{latexsym}
\usepackage{framed}
\usepackage{amsmath} 
\usepackage{amssymb}
\usepackage{graphicx}
\usepackage{url}
\usepackage{tabularx}
\usepackage{booktabs}
\usepackage{multirow}
\usepackage{soul}
\usepackage{subcaption}

\usepackage[T1]{fontenc}

\usepackage[utf8]{inputenc}
\usepackage{microtype}

\newcommand*{\affaddr}[1]{#1}

\title{Near-Negative Distinction:\\ Giving a Second Life to Human Evaluation Datasets}

\author{
  \quad \textbf{Philippe Laban}
  \quad \textbf{Chien-Sheng Wu}
   \quad \textbf{Wenhao Liu}
  \quad \textbf{Caiming Xiong} \\
  \affaddr{Salesforce AI Research} \\
  \{plaban, wu.jason, wenhao.liu, cxiong\}@salesforce.com
}

\begin{document}
\maketitle
\begin{abstract}
Precisely assessing the progress in natural language generation (NLG) tasks is challenging, and human evaluation to establish a preference in a model's output over another is often necessary.
However, human evaluation is usually costly, difficult to reproduce, and non-reusable.
In this paper, we propose a new and simple automatic evaluation method for NLG called Near-Negative Distinction (NND) that repurposes prior human annotations into NND tests.
In an NND test, an NLG model must place a higher likelihood on a high-quality output candidate than on a near-negative candidate with a known error.
Model performance is established by the number of NND tests a model passes, as well as the distribution over task-specific errors the model fails on.
Through experiments on three NLG tasks (question generation, question answering, and summarization), we show that NND achieves a higher correlation with human judgments than standard NLG evaluation metrics. We then illustrate NND evaluation in four practical scenarios, for example performing fine-grain model analysis, or studying model training dynamics. Our findings suggest that NND can give a second life to human annotations and provide low-cost NLG evaluation.
\end{abstract}

\section{Introduction}

\begin{figure}
    \centering
    \includegraphics[width=0.34\textwidth]{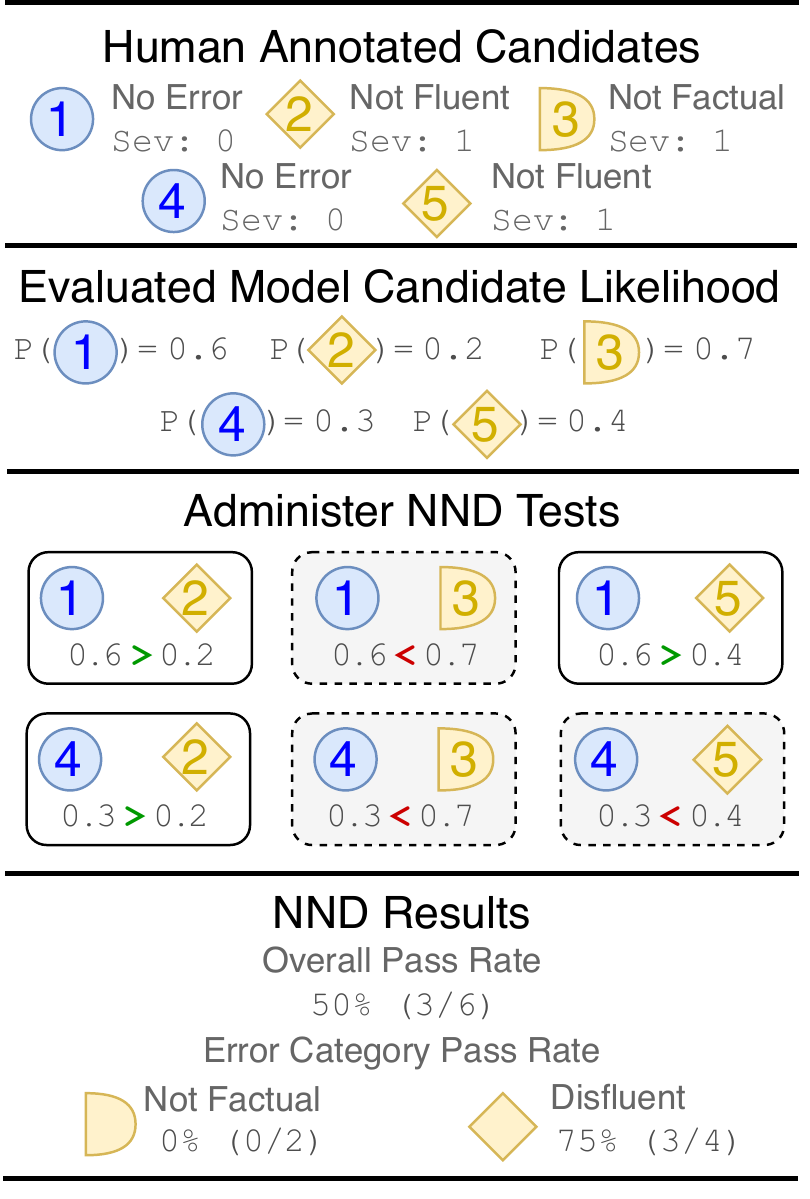}
    \caption{Stages of the \textbf{Near-Negative Distinction} framework for NLG evaluation. A pre-existing human evaluation is repurposed into a series of NND tests.}
    \label{figure:nnd_explainer}
\end{figure}

Pre-training of large language models has fueled recent progress in many natural language generation (NLG) tasks such as summarization \cite{zhang2020pegasus}, question answering \cite{tafjord2021general} \cite{ng2019facebook}, and question generation \cite{murakhovska2022mixqg}. However, quantifying this progress remains a challenge due to the open-ended nature of NLG.

The gold standard for NLG evaluation is manual expert annotation: it can be highly precise and fully customized to an NLG task, helping identify model limitations, and setting the direction of future work. The main limitation of manual expert annotation is the complexity and cost associated with running an evaluation. The cost often increases linearly or quadratically with the number of models compared, restricting evaluation to a small number of models.

To circumvent the cost of expert evaluation, many in the field rely on automatic metrics such as BLEU \cite{papineni2002bleu}, which compute text quality based on n-gram overlap between model outputs and human references. Such metrics are easy to compute, and have been shown to moderately correlate with human judgments, but are limited in three ways: they only offer an aggregate score that is difficult to interpret, they do not offer a clear upper bound in performance, and they have limited generalized ability to some NLG tasks \cite{liu-etal-2016-evaluate,sulem-etal-2018-bleu}.

In this paper, we propose a new and simple automatic framework for the evaluation of NLG models which we call Near-Negative Distinction (NND).
At a high level, the NND framework bridges the gap between expert annotation and automated metrics by repurposing existing annotations into a series of automatic tests which assess how likely a model is to avoid previously annotated errors.

The first contribution of our work is the definition of the NND framework, illustrated in Figure~\ref{figure:nnd_explainer}. In NND, an existing human evaluation dataset $D$ is repurposed to create a series of automated tests. For a given input context, $D$ should contain annotations for several model outputs, some high-quality (candidates 1, 4) and some low-quality (candidates 2, 3, 5). A collection of NND tests is created containing candidate pairs of differing quality. Generation models pass an NND test if they assign a higher likelihood to the high-quality candidate than the low-quality one. NND evaluation produces an overall test pass rate, as well as a pass rate for each error category, which can be used to inspect model strengths and weaknesses.

The second contribution is the creation of NND datasets from existing human evaluations for three NLG tasks: question generation, generative question answering, and summarization. On these three tasks, verification experiments find that NND pass rates correlate better with human judgments than existing evaluation metrics, both n-gram-based metrics such as BLEU \cite{papineni2002bleu}, and more recent metrics such as BERTScore \cite{zhang2019bertscore} and QuestEval \cite{zhang2019bertscore}.

The third contribution is a collection of practical experiments showcasing how to use NND. The experiments demonstrate the flexibility of the NND framework, showing it can be useful to extrapolate a model's performance in a user study, perform fine-grain model analysis, study scaling effects in model families, or discover trends during training.

Although we focus experiments on the English language, the NND framework is not English-specific, and we encourage the community to experiment with NND evaluation, helping to expand it to new NLG domains and languages.

We publicly release the NND datasets we generated as well as the code needed to create new NND datasets, and models used in experiments\footnote{\url{https://github.com/Salesforce/nnd_evaluation}}.

\section{Near-Negative Distinction}

We now detail the process of transforming pre-existing human annotations into an NND dataset and show how to perform NND evaluation.

\subsection{NND Dataset Creation Procedure}
\label{sec:nnd_creation_procedure}

A human annotation dataset $D$ consists of \texttt{(context, candidate)} tuples that have been annotated typically with one or more labels from a discrete error categorization. Several properties are required from human annotation datasets to be compatible with the NND framework.
First, several candidates should be annotated for each context, so that pairs of candidates can be formed into unit NND tests.
Second, it should be possible to map error categories to varying quality levels. For instance in Figure~\ref{figure:nnd_explainer}, candidate 1 labeled \texttt{No Error} is of higher quality than candidate 2 labeled \texttt{Not Fluent}. If these properties are present in a human annotation dataset, an NND dataset can be created in three steps:
\begin{enumerate}
    \item \textbf{Group By Context:} Group all annotated candidates for a given context, typically each candidate originates from an NLG model.
    \item \textbf{Assign Quality}: Assign a quality to each candidate within a group based on its annotation. 
    \item \textbf{Generate Candidate Pairs:} For a given context, construct all pairs of candidates of differing quality ($C_{high}$, $C_{low}$).
\end{enumerate}
The difference in quality between some error categories might not be known (e.g., the difference between ``Not Fluent'' and ``Not Factual'' candidates in Figure~\ref{figure:nnd_explainer}), preventing the ability to fully rank candidates. Because of this limitation, NND focuses on pairwise comparisons rather than ranking, analyzing each pair of candidates for which a quality differential is known.

\subsection{Administering NND}

The finalized NND dataset consists of \texttt{(context, $C_{high}$, $C_{low}$)} triplets we call \textit{NND tests}. Most text generators are language models, which assign a probability to a sequence of tokens. Sequence probability can be used during generation to rank partial candidates such as in beam search generation, however most often a generated sequence's likelihood is discarded once generation is completed.

In NND, we make use of sequence likelihoods to assess whether models are likely to reproduce the mistakes of previous models, or if they can correctly assign lower likelihood to low-quality candidates.
Formally, each candidate $C$ is tokenized into a sequence of tokens: $w_{1}, ... w_{N}$, and a candidate's likelihood is computed in the following way:

\begin{equation}
     LL(C) = \frac{\sum_{i=1}^{N} log (P(w_i \lvert \texttt{ct}, w_{1}, ..., w_{i-1}))}{N},
    \label{eq:likelihood}
\end{equation}
where $P(w_i \lvert ...)$ is the probability assigned by the model to the $i$-th token of the candidate, and \texttt{ct} is the input context. We use log likelihood instead of likelihood, a standard step to improve numerical stability. We further choose to normalize the likelihood by the sequence length ($N$) to counterbalance the effect of sequence length on likelihood.
An NND test is performed by computing the likelihood of both candidates $LL(C_{high})$ and $LL(C_{low})$ and comparing both. The model passes the test if
\begin{equation}
    LL(C_{high}) > LL(C_{low}).
\end{equation}
In cases where the model fails the test, the error category of $C_{low}$ is recorded, allowing to compute NND pass rates for each category of error.

By administering an entire dataset of tests, the NND produces two outputs: first an overall pass rate which is the percentage of NND tests passed by the model, and the breakdown of pass rates for each error category. The two outputs complement each other: the former can be used to compare models, and the second can be used to inspect model performance and discover model limitations.

\subsection{Verification of NND Quality}
\label{section:verification_definition}

To gain an understanding of the quality of NND estimates, we run verification experiments assessing the level of correlation between NND estimates of model performance and human reference annotations. We run identical verification experiments with a set of standard NLG metrics.

We design two verification experiments based on desired properties for an evaluation metric: (1) \textbf{Rank Correlation}, an evaluation metric should rank NLG models similarly to rankings based on human annotation, (2) \textbf{Gap Correlation}, a metric's estimate of gaps in performance between pairs of models should correlate positively with gaps measured through human annotation (i.e., if human annotation reveals a large gap in performance between two models, the evaluation metric should similarly estimate a large gap).

For Rank Correlation, given a set of NLG models and a metric, we compute the Kendall rank correlation coefficient ($\tau$)~\cite{kendall} between the models' ranking according to the metric, and the ranking based on human annotation. Higher $\tau$ signifies that an evaluation metric is more accurate at predicting the ordinal ranks of models.

For Gap Correlation, for each pair of NLG models, we compute the difference in performance according to the metric and according to human annotation. The gaps of all pairs of models are assembled into two vectors of size ${n \choose 2}$, and we compute the Pearson correlation of the two vectors. If a metric achieving a high Gap Correlation is well calibrated and can predict gaps in performance between two models accurately.

In Section~\ref{section:nnd_datasets}, we introduce NND datasets for three NLG tasks, based on pre-existing human annotations. In Section~\ref{section:nnd_verification}, we perform the verification experiments in the three domains and confirm that NND correlates better with human opinion than well-established NLG metrics. Section~\ref{section:nnd_applications} introduces practical use-cases of NND evaluation.

\section{NND Datasets}
\label{section:nnd_datasets}
\subsection{NND For Question Generation}
\label{sec:nnd_qgen}

We first describe NND experiments for the task of Question Generation, based on Quiz Design (QD) dataset \cite{laban2022quiz}. For each context in QD, seven answer-aware QGen models generated up to seven questions. Ten teachers designing educational quizzes annotated 3,164 questions with one of four error types: \textit{No Error, Disfluent, Off Target, Wrong Context}.

We generate NND tests by pairing \textit{No Error} questions with any question with an error, producing 2,686 NND pairs in total. Examples in Table~\ref{appendix:nnd_ex_qgen}.


We run NND experiments with the seven models used in the original QD study (GPT2-\{distil,base,med\} \cite{radfordlanguage}, BART-\{base,large\} \cite{lewisbart}, ProphetNet, and MixQG-Large \cite{murakhovska2022mixqg}), as well as three newer models that were not released when the QD annotation was run: MixQG-3B, and Macaw-\{3B-11B\} \cite{tafjord2021general}.

\subsection{NND For Question Answering}
\label{sec:nnd_qa}

In generative QA, a QA model receives a question and must generate a potentially abstractive answer. We create an NND dataset by re-purposing the Challenge 300 annotations \cite{tafjord2021general}. Challenge 300 is a suite of 300 questions intended to challenge QA models (e.g., Can you sit and stand at the same time?). For each question, QA models must generate a free-text answer, and candidate answers from five large QA models (including GPT3) were annotated with a credit of either 0 (incorrect), 0.5 (partially correct), or 1 (correct). Each question in Challenge 300 is further tagged into 20 categories, which we consolidate into 5 groups: Common Sense, Comparison, Entity, Creativity, and Science. We create NND test pairs out of (correct, incorrect) answer pairs and obtain 829 NND test pairs which we further organize according to category groups. Example NND tests for each category in Table~\ref{appendix:nnd_ex_qa}.

We run NND experiments with three families of publicly available generative QA models: T5 finetuned on Natural Questions \cite{roberts2020much}, UnifiedQA \cite{khashabi2020unifiedqa}, and Macaw \cite{tafjord2021general}, which achieved the highest performance during annotation.

\subsection{NND For Summarization}
\label{sec:nnd_summ}

For summarization, we adapt two human annotation datasets to the NND framework: SummEval \cite{fabbri2021summeval} and FRANK \cite{pagnoni2021understanding}. Example NND tests in Table~\ref{appendix:nnd_ex_summ}.

SummEval consists of 100 documents each with 8 to 9 system-generated summaries annotated with 5-Point Likert scale ratings on four general attributes (Consistency, Coherence, Fluency, and Relevance). We treat each attribute independently, and normalize Likert scale annotations following the SummaC benchmark procedure \cite{laban2021summac}: for each attribute, a summary is of high quality if a majority of annotators gave the summary a score of 5, and is of low quality otherwise. The NND procedure yields 3,613 NND tests.

FRANK focuses annotation on the consistency attribute, offering more specialized error categories. The test portion of FRANK contains 350 news articles, each coupled with 4 or 5 summaries. Each summary has annotations that follow a hierarchical error categorization, breaking down consistency errors into four groups: \textit{No Error, Semantic Frame, Discourse, and Verifiability} errors.\footnote{We remove the ``Other'' category as it has few samples.} We treat \textit{No Error} as high-quality, and any other error as low-quality, and generate 824 NND test pairs.

We run NND experiments with five summarization models in the SummEval evaluation (M9, M17, M20, M22, M23) and perform a fine-grain comparison of BART-large and PEGASUS  \cite{zhang2020pegasus}, two models that achieve very strong ROUGE performance on the CNN/DM dataset \cite{nallapati2016abstractive}.

\section{NND Verification}
\label{section:nnd_verification}
\begin{table}[]
    \resizebox{0.5\textwidth}{!}{%
    \begin{tabular}{lcccccc}
     & \multicolumn{2}{c}{\textbf{QGen}} & \multicolumn{2}{c}{\textbf{Gen. QA}} & \multicolumn{2}{c}{\textbf{Summ.}} \\
    \cmidrule(r){1-1} \cmidrule(l){2-3} \cmidrule(l){4-5} \cmidrule(l){6-7}
    \textbf{Metric} & \textbf{Rank} & \textbf{Gap} & \textbf{Rank} & \textbf{Gap} & \textbf{Rank} & \textbf{Gap} \\
     & $\tau$ & $r$ & $\tau$ & $r$ & $\tau$ & $r$ \\
    \cmidrule(r){1-1} \cmidrule(l){2-3} \cmidrule(l){4-5} \cmidrule(l){6-7}
    BLEU      & 0.65 & 0.35 & 0.40 & 0.63 & 0.45 & 0.74 \\
    R-1       & 0.64 & 0.31 & 0.40 & 0.57 & 0.55 & 0.84 \\
    R-2       & 0.65 & 0.34 & 0.27 & 0.56 & 0.55 & \textbf{0.86} \\
    R-L       & 0.65 & 0.41 & 0.40 & 0.57 & 0.55 & 0.85 \\
    METEOR    & 0.65 & 0.36 & 0.27 & 0.56 & 0.55 & 0.72 \\
    BERT & 0.49 & 0.36 & 0.40 & 0.57 & 0.65 & 0.67 \\
    BARTScore & 0.65 & 0.40 & 0.27 & 0.57 & \textbf{0.75} & 0.74 \\
    QuestEval & 0.65 & 0.39 & 0.27 & 0.56 & \textbf{0.75} & 0.79 \\
    NND       & \textbf{0.78} & \textbf{0.80} & \textbf{0.67} & \textbf{0.86} & 0.70 & \textbf{0.86} \\
    \bottomrule{}
    \end{tabular}
    }
    \caption{\textbf{Results for the Rank ($\tau$) and Gap ($r$) Correlation experiments.} Experiments were performed for Question Generation, Generative QA and Summarization using scores from standard NLG evaluation metrics and NND. Each entry is the average of verification experiments run on the dataset.}
    \label{table:verification}
\end{table}

We now present results from running the verification experiments of Section~\ref{section:verification_definition} on the three domains we study. In our analysis, we compare NND to standard n-gram based evaluation metrics: BLEU \cite{papineni2002bleu}, ROUGE \cite{lin2004rouge}, METEOR \cite{banerjee2005meteor}, as well as more recent Transformer-based metrics: BERTScore \cite{zhang2019bertscore}, BARTScore \cite{yuan2021bartscore} and QuestEval \cite{scialom2021questeval}. For each verification experiment, we are limited to evaluating models present in the annotation datasets that have been open-sourced as the NND framework requires a running version of the model to compute candidate likelihoods.

For QGen, verification experiments used all seven models present in the annotations dataset, with separate verification experiments run on each of the three error types.
For QA, verification experiments involved three of the four available models\footnote{The fourth model GPT3-DaVinci is not publicly released}, and were run on each question category.
For Summarization, verification experiments were run with five summarizers from SummEval (M9, M17, M20, M22, M23) with experiments run on each of the four summarization aspects. We do not run verification experiments on FRANK, as it contains fewer annotations of publicly released models.

Verification results summarized in Table~\ref{table:verification}. NND compares favorably across the board, achieving the highest correlation on five of the six assessments. Improvements in correlation are stronger on the QG and generative QA tasks than Summarization, on which ROUGE, BARTScore, and QuestEval achieve strong performance.

We note an important conceptual difference between NND and the metrics we compare to which are reference-based. Reference-based metrics score a generator by establishing a similarity between the model's candidate outputs and human-written references. In contrast, NND is reference-less and relies on human annotations of several model candidate outputs to evaluate models. We hypothesize that the use of near-negatives, and whether a model is likely to avoid them, provides a useful signal that leads to high-quality model evaluation.

We next turn to use the NND framework in practical situations and assume that NND pass rates provide quality estimates of model ranks and performance gaps between models.

\section{NND Applications}
\label{section:nnd_applications}


\subsection{Extrapolating Model Performance}
\label{section:extrapolating}
\begin{table}[!htbp]
    \centering
    \resizebox{0.49\textwidth}{!}{%
    \begin{tabular}{lccccccccc}
     & & \multicolumn{4}{c}{\textbf{Quiz Design NND}} \\
    \cmidrule(r){3-6}
     & & \textbf{Overall} & \textbf{Disfluent} & \textbf{Off Tgt} & \textbf{W. Ctxt} \\
    \cmidrule(r){3-6}
    \multicolumn{2}{l}{\#NND Tests} & 2686 & 711 & 890 & 1085 \\
    \toprule{}
    
    \textbf{Model} & \textbf{\%A} & \multicolumn{4}{c}{\textbf{NND Test Pass Rate (\%)}} \\
    \cmidrule(r){1-2} \cmidrule(r){3-6}

    Distil-GPT2 & 33.4 & 44.9 & 52.7 & 37.0 & 46.0 \\
    GPT2-base   & 40.9 & 52.3 & 60.3 & 49.7 & 49.3 \\
    GPT2-med    & 51.3 & 60.8 & 63.3 & 64.5 & 56.1 \\
    BART-Base   & 52.0 & 59.6 & 60.5 & 64.5 & 55.0 \\
    ProphetNet  & 53.5 & 67.7 & 58.1 & 79.8 & 64.1 \\
    BART-Large  & 58.4 & 64.2 & 63.3 & 70.8 & 59.4 \\
    MixQG-L & 68.4 & 70.9 & 66.9 & 80.9 & 65.3 \\
    \cmidrule(r){1-2} \cmidrule(r){3-6}
    MixQG-3B    & - & \textbf{72.9} & 69.5 & \textbf{81.7} & \textbf{67.8} \\
    Macaw-3B    & - & 69.2 & \textbf{70.3} & 73.3 & 65.1 \\
    Macaw-11B   & - & 70.6 & 69.3 & 78.0 & 65.4 \\
    \bottomrule{}
    \end{tabular}
    }
    \caption{\textbf{Extrapolation of QGen model's performance on the Quiz Design manual evaluation.} The first seven models (top) are part of the human evaluation (\%A: original human acceptance rate), bottom three are only evaluated with NND.}
    \label{table:qgen_results}
\end{table}

In Quiz Design, the largest MixQG-3B model was not included in the annotations due to latency requirements for the interface \cite{laban2022quiz}. Further, new QGen models have been released since the study's conduct. We leverage NND's ability to provide category-specific estimates of performance to extrapolate how these unseen models would have performed in the Quiz Design Study.

We run NND experiments for each of the seven models included in the study, as well as the unseen models. Results are summarized in Table~\ref{table:qgen_results}.

First, the three novel models all achieve strong performances, obtaining three of the best four overall NND pass rates. The MixQG-3B achieves the highest performance overall, seeing a total improvement of 2\% when compared to MixQG-L, the best performer at the time of the study, with gains on all three error categories. The Macaw models achieve the strongest performance in \textit{Disfluency}, but lower performance on \textit{Off Target} and \textit{Wrong Context} lead to lower performance overall.

These results show that NND can be used to give a second life to human evaluation datasets by projecting model performance a posteriori.

\subsection{Fine-Grained Model Comparison}
\label{section:bart_v_pegasus}

\begin{figure}
    \centering
    \begin{subfigure}[b]{0.48\textwidth}
        \centering
        \includegraphics[width=\textwidth]{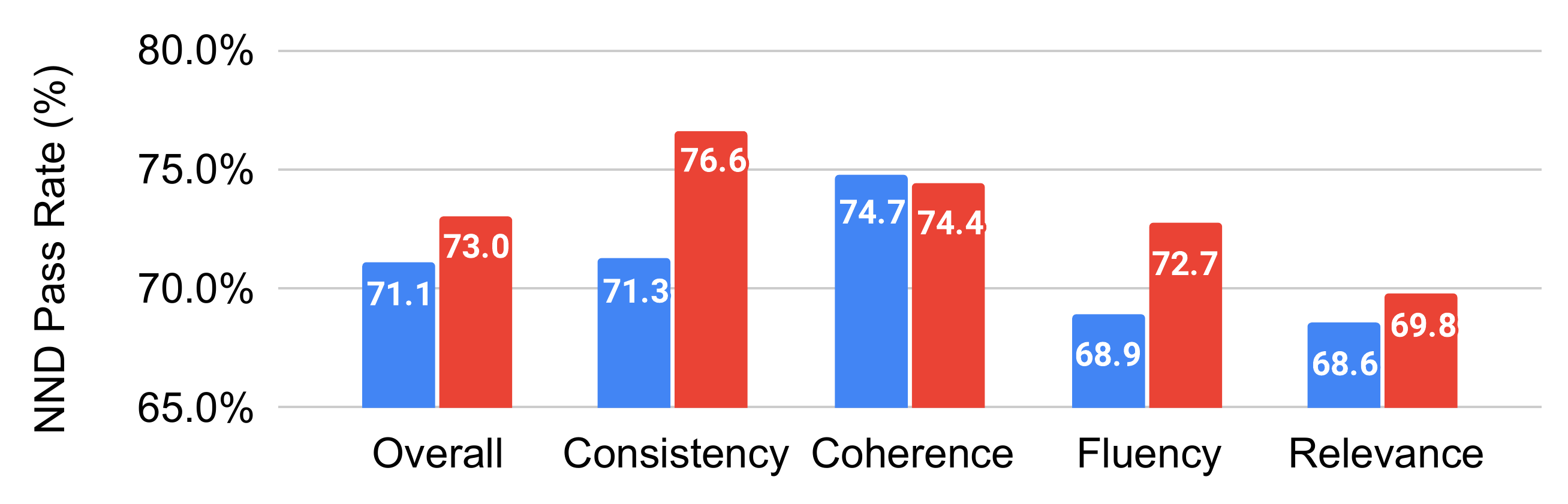}
        \caption{SummEval NND}
    \end{subfigure}

    \vskip\baselineskip
    \begin{subfigure}[b]{0.48\textwidth}   
        \centering 
        \includegraphics[width=\textwidth]{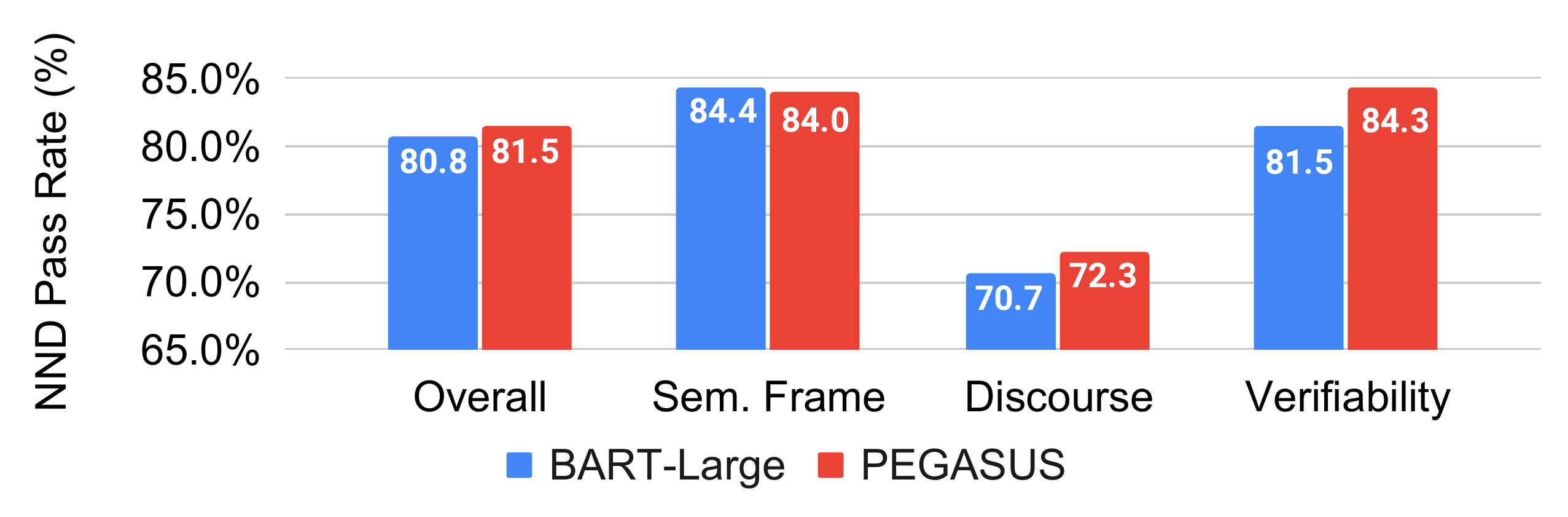}
        \caption{FRANK NND}
    \end{subfigure}
    \caption{\textbf{Fine-grain comparison of a pair of summarization models on based on two NND test sets.} SummEval estimates performance on four general aspects, and FRANK focuses on factual consistency errors.}
    \label{figure:summ_compare}
\end{figure}

Prior work has recognized the BART-Large and PEGASUS models as close contenders for top performance in summarization \cite{fabbri2021summeval}. The two models are virtually tied in terms of ROUGE-1 score on the CNN/DM test set with a variation of fewer than 0.1 points.

To gain specific insights into the differences between the models, we run NND experiments with both models using the general NND test set based on the SummEval annotations, as well as the factual consistency-focused FRANK annotations. Results are summarized in Figure~\ref{figure:summ_compare}.

On the SummEval test set, PEGASUS narrowly outperforms BART overall, owing to 4-5\% gains in the consistency and fluency aspects. Performance on the coherence and relevance aspects are narrower, with BART topping coherence, and PEGASUS with a slight edge in relevance.

The SummEval results are reaffirmed by the FRANK NND experiment, on which PEGASUS also outperforms BART overall, confirming that PEGASUS is better at avoiding factual errors than BART. However, on this more precise error categorization, PEGASUS does not win out entirely, with BART-Large achieving a higher pass rate on the Semantic Frame errors.

The NND results confirm that the two models' performance is close, with overall NND pass rates within 2\% of each other, yet reveal some subtlety in the specific strengths and weaknesses of each model. Depending on the application, certain attributes might be of more or less importance, and NND could inform a user on which model to select.

\subsection{Model Scaling Effects}

\begin{figure}
    \centering
    \begin{subfigure}[b]{0.20\textwidth}
        \centering
        \includegraphics[width=\textwidth]{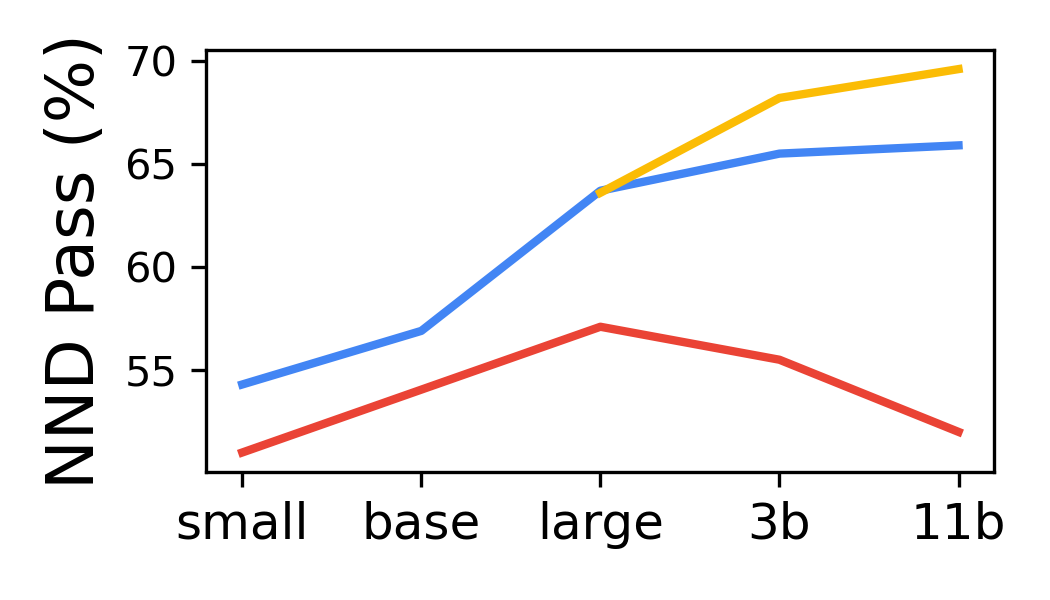}
        \caption{Overall}
    \end{subfigure}
    \hfill
    \begin{subfigure}[b]{0.27\textwidth}  
        \centering 
        \includegraphics[width=\textwidth]{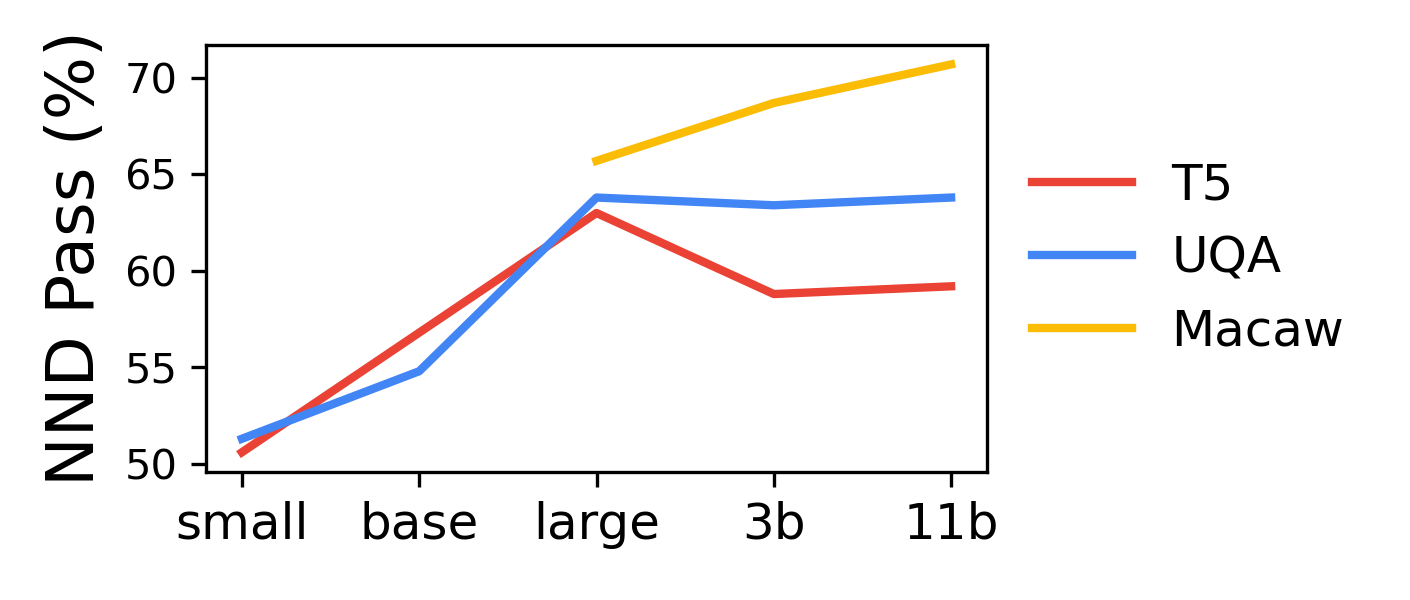}
        \caption{Common Sense}
    \end{subfigure}
    \vskip\baselineskip
    \begin{subfigure}[b]{0.20\textwidth}
        \centering 
        \includegraphics[width=\textwidth]{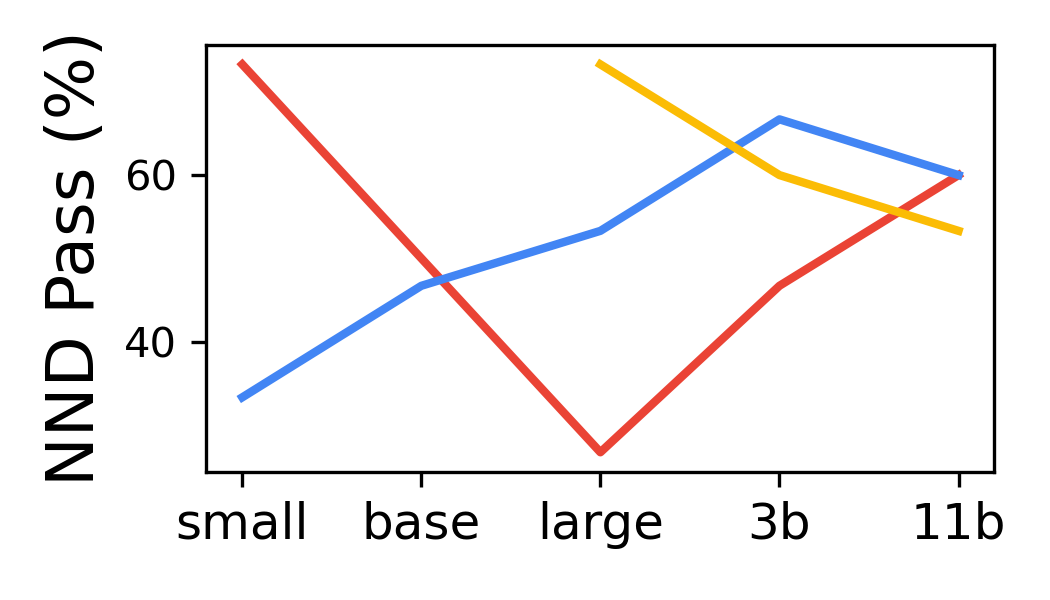}
        \caption{Comprehension}
    \end{subfigure}
    \hfill
    \begin{subfigure}[b]{0.27\textwidth}   
        \centering 
        \includegraphics[width=\textwidth]{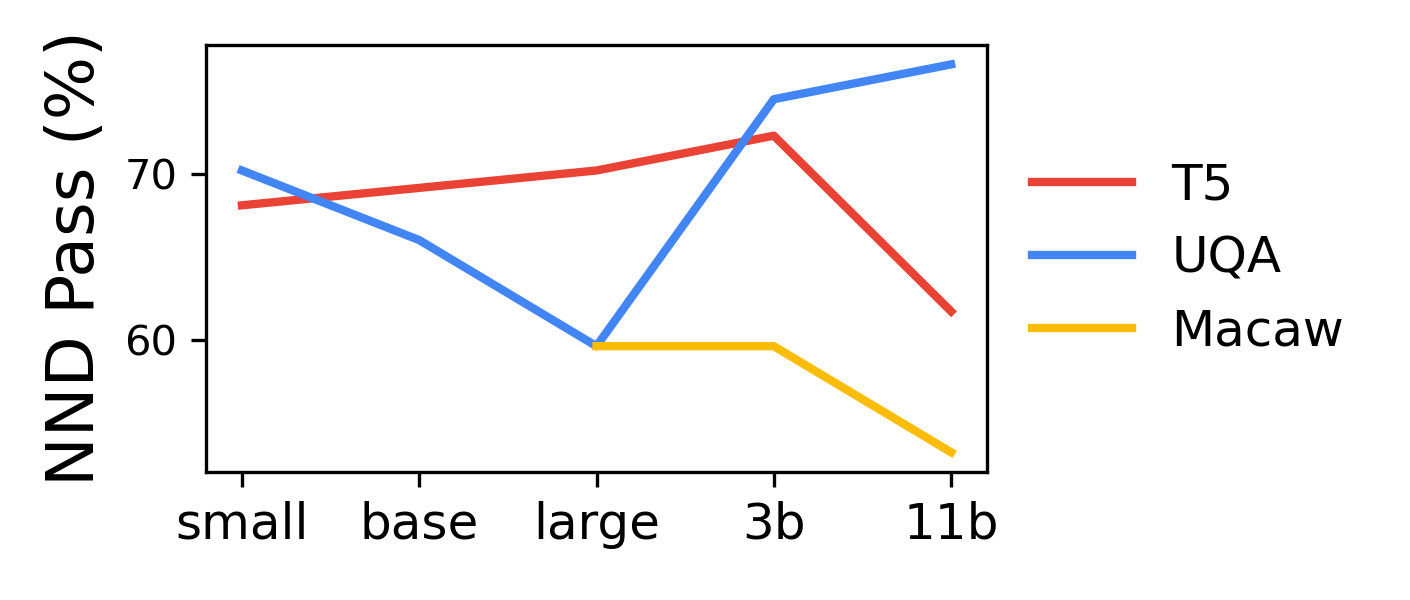}
        \caption{Entity}
    \end{subfigure}
    \vskip\baselineskip
    \begin{subfigure}[b]{0.20\textwidth}   
        \centering 
        \includegraphics[width=\textwidth]{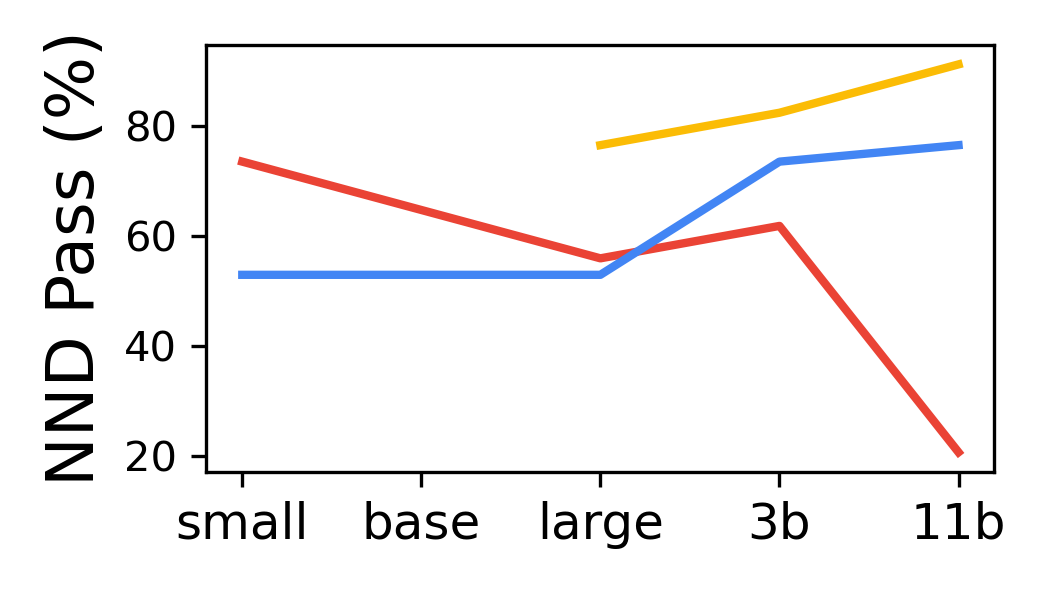}
        \caption{Creativity}
    \end{subfigure}
    \hfill
    \begin{subfigure}[b]{0.27\textwidth}   
        \centering 
        \includegraphics[width=\textwidth]{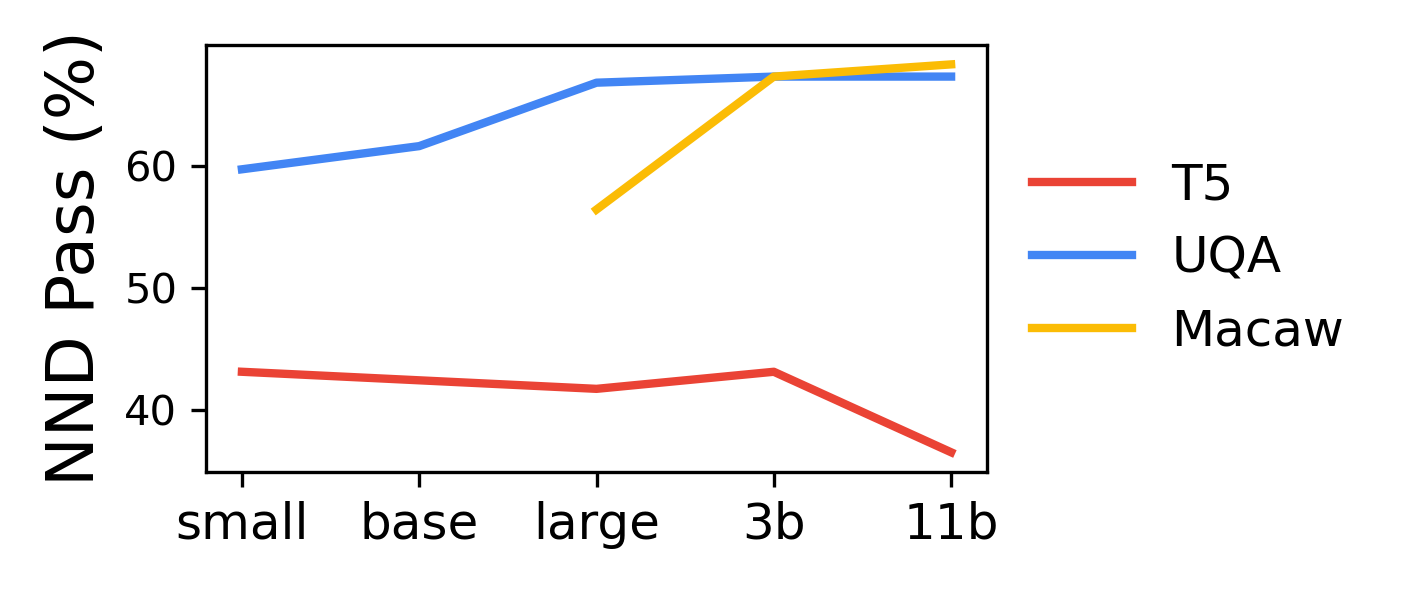}
        \caption{Science}
    \end{subfigure}

    \caption{\textbf{Scaling experimental results for QA models.} Overall and category-specific NND pass rates are computed for varying model sizes from three model families: T5, UnifiedQA and Macaw.}
    \label{figure:nnd_qa_scaling}
\end{figure}

The authors of the Challenge 300 dataset only annotated text outputs from the largest models available for each model family \cite{tafjord2021general}. This annotation strategy is understandable, as annotating smaller models' answers increases annotation cost, but it limits understanding of the effect of model size on performance.

We run NND experiments for all model sizes available for three families of QA models: T5 finetuned on Natural Questions (Small, Large, 3B, 11B) \cite{roberts2020much}, Unified-QA (Small, Base, Large, 3B, 11B) \cite{khashabi2020unifiedqa} and Macaw (Large, 3B, 11B) \cite{tafjord2021general}, with results summarized in Figure~\ref{figure:nnd_qa_scaling}.

Overall, increasing model size leads to gradual increases in performance for the UnifiedQA and Macaw models. Unexpectedly for T5, performance peaks with the T5-Large, however overall the T5 family underperforms UnifiedQA and Macaw.

Focusing on UnifiedQA and Macaw, model performance increases steadily in three question categories: Common Sense, Creativity, and Science, but surprisingly stagnates or decreases in the Comprehension and Entity categories.

The NND experiments reveal that although performance tends to improve with model size increase, the trends vary widely by question category: an end-user with a particular question category in mind might benefit from a smaller model size.

\subsection{Evaluation During Training}
\label{section:nnd_during_training}

\begin{figure}
    \centering
    \begin{subfigure}[b]{0.23\textwidth}
        \centering
        \includegraphics[width=\textwidth]{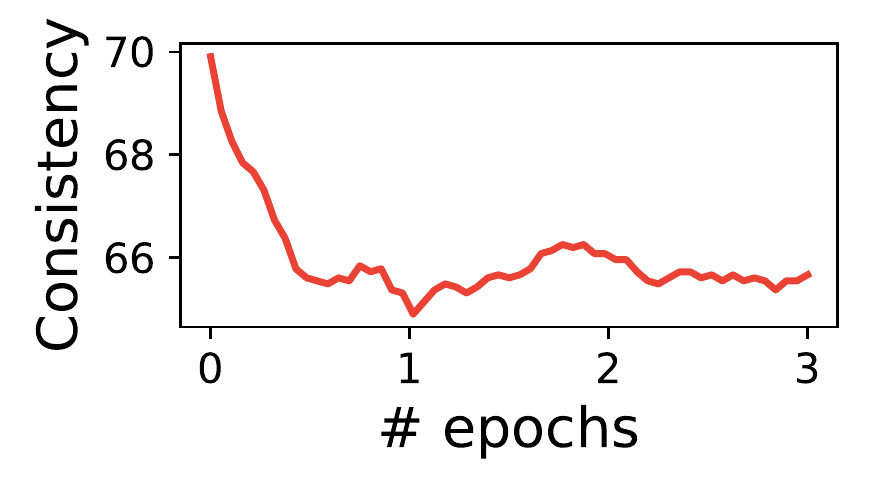}
    \end{subfigure}
    \hfill
    \begin{subfigure}[b]{0.23\textwidth}  
        \centering 
        \includegraphics[width=\textwidth]{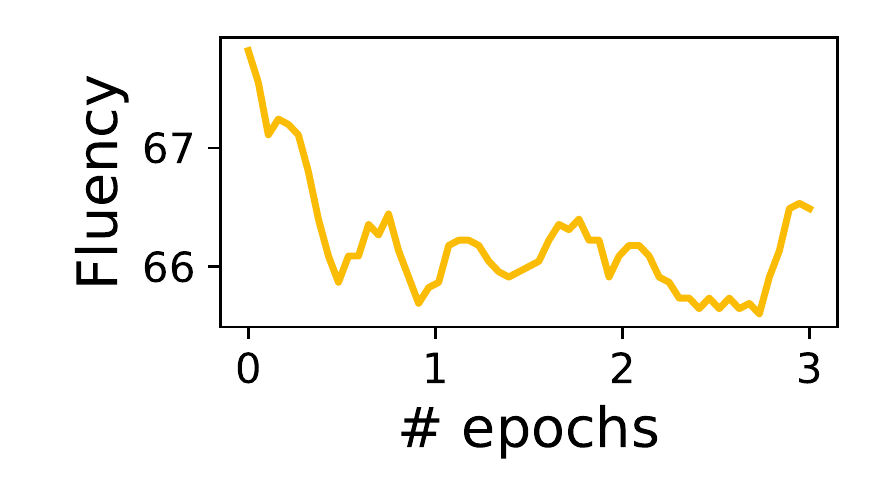}
    \end{subfigure}
    \vskip\baselineskip
    \begin{subfigure}[b]{0.23\textwidth}   
        \centering 
        \includegraphics[width=\textwidth]{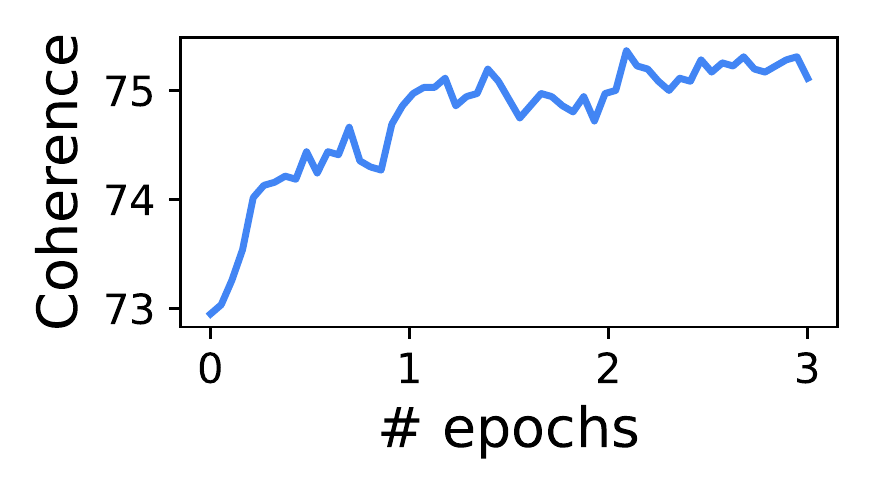}
    \end{subfigure}
    \hfill
    \begin{subfigure}[b]{0.23\textwidth}   
        \centering 
        \includegraphics[width=\textwidth]{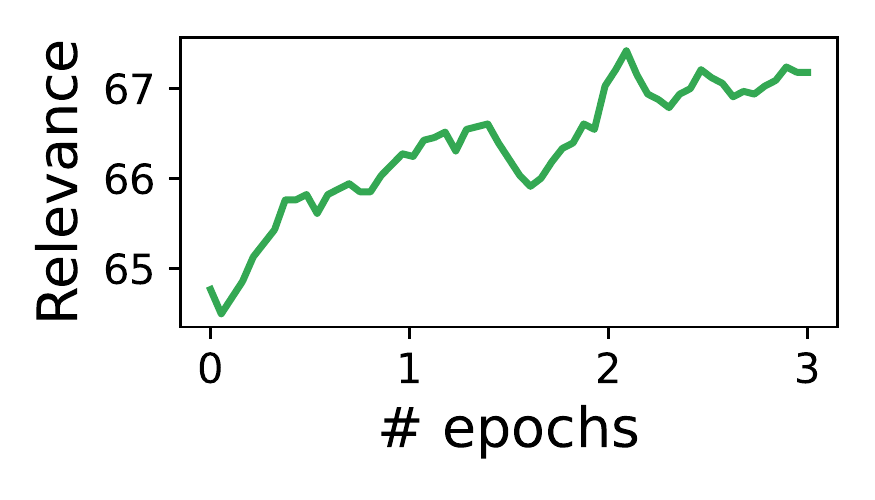}
    \end{subfigure}
    \caption{\textbf{NND Performance as a summarization model is trained.} By running SummEval NND evaluations on model checkpoints during training, model ability to detect consistency, fluency, coherence and relevance errors can be studied.}
    \label{figure:training_plots}
\end{figure}

So far, we ran NND to evaluate finalized models, performing comparisons across models. We now use NND to inspect a model during training.

We train a BART-base model on the CNN/DM dataset using teacher forcing with cross-entropy loss for three epochs. We perform an NND evaluation of the latest model checkpoint every 2,000 training steps, using the SummEval NND test pairs.

Results summarized in Figure~\ref{figure:training_plots}. 
Surprisingly, the model's ability to detect consistency and fluency errors decreases during training, with NND pass rates decreasing by 2-4\%. This finding mirrors the analysis of training dynamics in summarization, which finds that models become less factual in later stages of the training process \cite{goyal2021training}. On the other hand, model performance on coherence and relevance errors steadily increases during training.
These trends could be explained by the model becoming better at summarization-specific skills, such as content selection (relevance) and ordering (coherence) at the cost of factual consistency and general fluency.

\section{Related Work}

\textbf{NLG Benchmarks.} Following the success of benchmarks such as GLUE \cite{wang2018glue} for the evaluation of NLU models, some work has proposed benchmarks as a way to evaluate NLG models, such as GLGE \cite{liu2021glge} with 8 NLG tasks or the crowd-sourced BigBench \cite{bigbench2021} with 209 NLG tasks. More recently, the GEM Workshop proposed the GEM Benchmark \cite{gehrmann2021gem}, a living benchmark with rule-based challenge sets which can be updated with new models and reference-based metrics. Benchmarks are useful for broad comparison of model performance across tasks, for example with the evaluation of large language models in few-shot settings. We view NND as complementary to NLG benchmarks: a highly task-specific tool that can be used to assess a model's potential limitation on a particular task.

\textbf{LM Likelihood Score.} Language-modeling likelihood and perplexity (the exponentiation of log-likelihood) are commonly used to evaluate NLG models \cite{hashimoto2019unifying}. For example, test-set perplexity is the standard metric to compare unconditional language models \cite{chelba2013one,khandelwal2019generalization}.
Model capacity and vocabulary size affect likelihoods, and careful normalization is required for model-to-model comparisons \cite{jelinek1977perplexity}.
In NND, likelihoods are not compared across models, circumventing normalization needs. Furthermore, likelihood and perplexity lack interpretability, whereas NND mirrors error categories of human evaluations.

\textbf{External LM Likelihood.} Besides the evaluated model's own likelihood, some work has used an external language model's likelihood for scoring. BARTScore \cite{yuan2021bartscore} uses a BART model's likelihood to evaluate generated texts on faithfulness, precision, and recall factors. \citet{salazar2020masked} propose Masked-Language Model Scoring to repurpose BERT-style NLU models into producing pseudo-log likelihoods shown to measure textual fluency. Although large external language models can be useful for measuring general language quality, it is challenging for a single model to assess the task-specific quality of generated text. In NND, test pairs are targeted at evaluating model performance on specific task skills.

\textbf{Contrastive Learning.} The use of negative candidates in NLG has been explored with recent interest in applying contrastive learning \cite{chopra2005learning} methods to NLG training \cite{he2020negative,liu2021simcls, Cao2021CLIFFCL}. In contrastive learning, a model being trained receives both positive and negative candidates and has a two-sided objective of increasing the likelihood of positive candidates, while reducing the likelihood of negative candidates.

Similarly, \textbf{Self-Critical Sequence Training} \cite{rennie2017self,Laban2021KeepIS} is an RL training method in which models generate several candidates which are scored and contrasted. NND relies on pairs of candidates of differing quality as well, however, the framework is focused on evaluation and not training. Further, SCST relies on automatic metrics to score negative candidates, whereas NND is based on human annotations. When a large number of NND tests are available, NND could be compatible with contrastive learning: a portion of the tests can be for model training, while a portion is reserved for evaluation.

\textbf{Language Model Behavioral Analysis.} Recent work has built behavioral analysis corpora \cite{isabelle2017challenge, naik2018stress, vig2020causal} to evaluate model behavior and bias. For example, in: ``The nurse said that \_ is fine'', a biased model assigns a higher likelihood to a stereotypical ``she'' pronoun than an anti-stereotypical pronoun (``he'', ``it''). Behavioral analysis corpora rely on unit tests, and models are evaluated by the percentage of passed tests. Unlike NND, behavioral analysis often relies on rules or a lexicon to construct tests and is focuses on the effect of a single word or phrase, whereas NND relies on model-generated candidates with human annotations.

\textbf{Datasets Repurposing} is common in machine learning and NLP \cite{koch2021reduced, koesten2020dataset}, particularly in cases where data access is limited or noisy. Common datasets, such as the Penn Treebank for syntax parsing \cite{marcinkiewicz1994building}, CNN/DM for summarization\cite{nallapati2016abstractive}, or PPDB for paraphrase detection and generation\cite{pavlick2015ppdb}. However, there are known limitations to fixed leaderboards, and some work has proposed evolving evaluation sets to accompany model improvements \cite{ma2021dynaboard,khashabi2021genie,kasai2021bidimensional}. With NND evaluation, we propose to repurpose the annotations of model-generated texts, both enabling to learn from prior model's errors, as well as adapt to more recent model performance.

\section{Discussion}

\subsection{Other Domains}
\label{sec:other_domains}

Although we focus on three NLG tasks, annotations from human evaluation in other NLG tasks could be used to expand the framework further in future work, for example with the WMT MQM \cite{freitag2021experts} annotations for translation, the SAMSA \cite{sulem2018semantic} annotations for text simplification, or the HLGD for news headline generation \cite{laban2021news}.

\subsection{Benefits of NND}

\textbf{Flexibility of Framework.} NND relies on pre-existing human annotations to generate NND test pairs. However, the required annotation format is flexible, our experiments show that NND is compatible with single-error categorizations (e.g., the Quiz Design in Section~\ref{sec:nnd_qgen}), hierarchical categorizations (e.g., FRANK in Section~\ref{sec:nnd_summ}), or Likert-scale ratings (e.g., SummEval in Section~\ref{sec:nnd_summ}). NND results adopt the shape of the repurposed human evaluation, for instance, results in Section~\ref{section:bart_v_pegasus} are broken down both by general summarization aspects using the SummEval NND, and further refined to detailed categories with the FRANK NND.

\textbf{Direct Language Model Evaluation.} In a typical NLG evaluation, a decoding strategy is used to generate a candidate which is evaluated. Often, authors of a model recommend a decoding strategy to pair with the model, which creates an additional confounding factor in the evaluation, as a better decoding strategy (e.g., Nucleus Sampling \citet{holtzman2019curious}) can lead to improvements regardless of model quality. NND avoids this problem by evaluating a model directly through its likelihood and by-passing the use of a decoding strategy.

\textbf{Computationally Inexpensive.} Computing candidate likelihood requires a single model forward pass, through teacher forcing, whereas other automated NLG evaluations often require full candidate generation, which is computationally expensive. The low computational cost of NND enables rapid evaluation during training (Section~\ref{section:nnd_during_training}).

Limitations of NND are discussed in Section~\ref{section:limitations}.

\section{Conclusion}

We introduce the Near-Negative Distinction (NND) framework for the evaluation of NLG models. In the NND framework, a pre-existing human evaluation dataset is repurposed to create NND test pairs comprised of text candidates of differing quality. Models are evaluated on their ability to assign a higher likelihood to high-quality candidates, giving an estimate of whether models would avoid the errors of previously evaluated models.
We apply the NND framework to three NLG tasks: question generation, question answering, and summarization, and show that NND results correlate better with human preference than prior NLG evaluation methods.
The NND framework allows the breaking down of model performance by error category, and we illustrate how the framework's flexibility can be used to understand model strengths and weaknesses, for instance extrapolating how newer models would perform in an existing human study or how a summarization model can lose factual consistency ability during training. NND is a simple, automatic, and versatile evaluation method that we hope can accelerate NLG research.

\section{Limitations}
\label{section:limitations}
\textbf{Reliance on Likelihood.} Not all NLG models are language models capable of producing candidate likelihoods. For instance, black-box models such as GPT-3 \cite{brown2020language} or an extractive summarizer \cite{mihalcea2004textrank} cannot be evaluated through NND out-of-the-box as there is no way to administer NND tests. Furthermore, NND relies on models being well-calibrated. If a model is poorly calibrated, it could generate a single quality candidate, but a poor judge of quality on other candidates, leading to low performance on NND tests. However, prior work has argued that model calibration is important: it enables models to generate diverse candidates and is important in gaining a user's trust in practical applications \cite{guo2017calibration}.

\textbf{Reliance on Prior Errors.} NND relies on annotated errors of previous models to evaluate a new model, which assumes errors made by models remain constant over time. This is limiting, as each generation of models has specific strengths and weaknesses, with new categories of errors emerging over time. We recommend that NND be used as a temporary extension to a human evaluation, allowing for a few generations of models to be evaluated on the same benchmark. However, the gold standard of NLG evaluation remains human evaluation, and it should still be performed frequently, and repurposed into updated NND test sets.

\textbf{NND Requirements.} Not all human annotations of generated texts can be repurposed for NND evaluation, and the two requirements -- outlined in \S\ref{sec:nnd_creation_procedure}) -- limit usability of the evaluation methodology. More precisely, annotations can be repurposed only if several model outputs are labeled for a given input, and if a partial ordering of quality over the labels is known. We however show in the paper that these requirements are common amongst existing annotation collections.

\textbf{Sensitivity to Normalization.} A complication of the NND framework is that it relies on inputting the prior model's outputs into the evaluated model to obtain a likelihood. NLG models use different norms for punctuation and capitalization, making the exchange of generated text across models delicate. Other NLG evaluation metrics are also sensitive to un-normalized texts \cite{post2018call}, and for NND it falls on the creator of the dataset to verify that NND test pairs are well-framed and do not contain noise that might affect result validity.

\section{Ethical Considerations}

We focused our experiments on models and datasets for the English language, and even though we expect the NND framework to be adaptable to other languages and settings, we have not verified this assumption experimentally and limit our claims to the English language.

The models and datasets utilized primarily reflect the culture of the English-speaking populace. Gender, age, race, and other socio-economic biases may exist in the dataset, and models trained on these datasets may propagate these biases. Question-answering and summarization tasks in particular have previously been shown to contain these biases.

We selected question generation, question answering, and summarization as the three NLG domains on which we assessed the NND framework. We expect that the framework will be beneficial in other NLG tasks such as data-to-text, image captioning, or simplification, but have not created NND test sets for these domains and limit our claims to the three tasks we ran experiments for.

We note that NND datasets are not novel datasets. Still, transformations of pre-existing human annotation datasets and proper permission to reuse underlying datasets should be granted before usage in the NND framework. Our experiments all relied on publicly released human evaluation annotations with explicit permission for research re-use.

\section*{Acknowledgement}
We thank Alexander Fabbri, Jesse Vig, Greg Durrett, Jiacheng Xu and Shafiq Joty for helpful feedback on the manuscript.

\bibliography{anthology,custom}
\bibliographystyle{acl_natbib}

\appendix

\section*{Appendix}
\renewcommand{\thetable}{A\arabic{table}}
\setcounter{table}{0}
\renewcommand{\thefigure}{A\arabic{figure}}
\setcounter{figure}{0}

\section{NND Examples}

We provide example NND tests from each of the datasets used in experimentation, with question generation examples in Table~\ref{appendix:nnd_ex_qgen}, generative QA in Table~\ref{appendix:nnd_ex_qa}, summarization in Table~\ref{appendix:nnd_ex_summ}. The elements were hand-picked to illustrate a diversity of cases and error categories present in the NND test sets.

\begin{table*}[]
    \resizebox{\textwidth}{!}{%
    \begin{tabular}{p{8cm} p{8cm}}
    \multicolumn{2}{c}{\Large{Selected NND Tests - Question Generation}} \\
    \toprule
    \\
    \multicolumn{2}{c}{\parbox{16cm}{Like all catalysts, enzymes \textbf{increase the reaction rate by lowering its activation energy}. Some enzymes can make their conversion {[}...{]}}} \\
    \\
    \hspace{2.8cm} What do enzymes do? & \hspace{2.7cm} What does enzyme do? \\
    \hspace{3.5cm} \texttt{No Error} & \hspace{3.5cm} \texttt{Disfluent} \\

    \\
    \hline
    \\
    \multicolumn{2}{c}{\parbox{16cm}{Californium {[}...{]} The element was named after the \textbf{university and the U.S. state of California}. Two crystalline forms exist for californium [...]}} \\
    \\
    \hspace{1.7cm} What is Californium named after? & \hspace{2.1cm} What is the state of California? \\
    \hspace{3.5cm} \texttt{No Error} & \hspace{3.5cm} \texttt{Off Target} \\

    \\
    \hline
    \\
    \multicolumn{2}{c}{\parbox{16cm}{The Palazzo Pitti {[}...{]} Giorgio Vasari proposed that Brunelleschi was the palazzo's architect, and that his pupil \textbf{Luca Fancelli} was merely his assistant in the task, but today it is Fancelli who is generally credited.}} \\
    \\
    \hspace{1.7cm} Who is generally credited with the & \hspace{1.7cm} Who was the pupil of Brunelleschi? \\
    \hspace{2.2cm} design of the Palazzo Pitti? & \\
    \hspace{3.5cm} \texttt{No Error} & \hspace{3.2cm} \texttt{Wrong Context} \\

    \bottomrule
    \end{tabular}
    }
    \caption{\textbf{Three selected examples from the NND QGen dataset.} For a given context with the target answer in bold, two candidates are provided: No Error (left) and Error (right).}
    \label{appendix:nnd_ex_qgen}
\end{table*}

\begin{table*}[]
    \resizebox{\textwidth}{!}{%
    \begin{tabular}{p{8cm} p{8cm}}
    \multicolumn{2}{c}{\Large{Selected NND Tests - Generative Question Answering}} \\
    \toprule
    \\
    \multicolumn{2}{c}{\Large{Common Sense}} \\
    \multicolumn{2}{c}{If plastic was a conductor, then would a plastic spoon conduct electricity?} \\
    \\
    \hspace{0.5cm} Yes, but it would be a very poor conductor. & \hspace{0.5cm} No. Plastic is a non-conductor of electricity.  \\
    \hspace{2.75cm} \texttt{Credit: 1} & \hspace{2.75cm} \texttt{Credit: 0} \\
    \\
    \hline
    \\
    \multicolumn{2}{c}{\Large{Comparison}} \\
    \multicolumn{2}{c}{What is the difference between a noun and a verb?} \\
    \\
    a verb expresses action, a noun describes things & A noun is a person, place, or thing. A verb is a person, place, or thing. \\
    \hspace{2.75cm} \texttt{Credit: 1} & \hspace{2.75cm} \texttt{Credit: 0} \\
    \\
    \hline
    \\
    \multicolumn{2}{c}{\Large{Entity}} \\
    \multicolumn{2}{c}{\parbox{16cm}{Imagine an empty cup. Now put a coin in the cup. Now put another coin in the cup. Now put a pen in the cup. How many coins are in the cup now?}} \\
    \\
    \hspace{2.5cm} 2 coins and a pen. & \hspace{3.5cm} three \\
    \hspace{2.75cm} \texttt{Credit: 1} & \hspace{2.75cm} \texttt{Credit: 0} \\
    \\
    \hline
    \\
    \multicolumn{2}{c}{\Large{Creativity}} \\
    \multicolumn{2}{c}{How can you sit and stand at the same time?} \\
    \\
    \hspace{3.0cm} you can't & \hspace{1.0cm} It's easy. You just sit down and stand up. \\
    \hspace{2.75cm} \texttt{Credit: 1} & \hspace{2.75cm} \texttt{Credit: 0} \\
    \\
    \hline
    \\
    \multicolumn{2}{c}{\Large{Science}} \\
    \multicolumn{2}{c}{Why does the sky reflect blue light?} \\
    \\
    \hspace{0.25cm} The sky is blue because of Rayleigh scattering. & \hspace{2.0cm} Because God hates you.  \\
    \hspace{2.75cm} \texttt{Credit: 1} & \hspace{2.75cm} \texttt{Credit: 0} \\
    \\
    \hline
    \bottomrule
    \end{tabular}
    }
    \caption{\textbf{Three selected examples from the NND QGen dataset.} For a given context with the target answer in bold, two candidates are provided: No Error (left) and Error (right).}
    \label{appendix:nnd_ex_qa}
\end{table*}

\begin{table*}[]
    \resizebox{\textwidth}{!}{%

    \begin{tabular}{p{8cm} p{8cm}}
    \multicolumn{2}{c}{\Large{Selected NND Tests - Summarization}} \\
    \toprule \\
    
    \multicolumn{2}{c}{\parbox{16cm}{Uber has poached Facebook's security chief Joe Sullivan in an attempt to double down on rapidly escalating safety concerns. The \$40 billion taxi service has been plagued by serious accusations of failing to vet its drivers. Lawsuits have been brought against Uber in San Francisco and Los Angeles. A New Delhi driver was accused of raping a passenger in December. This week in Denver, a driver tried and failed to break into a passenger's home. And in London, [...]}} \\
    \\
    Facebook's security chief Joe Sullivan will leave his role as Facebook's security chief to help Uber defend safety concerns. Lawsuits have been brought against Uber in San Francisco and Los Angeles. There were three high-profile assault cases involving Uber drivers in December 2014. & The \$40 billion taxi service has been plagued by serious accusations. The \$40 billion taxi service has been plagued by serious accusations. It comes days after a driver tried and failed to break into a passenger's home.\\
    \hspace{3.5cm} \texttt{No Error} & \hspace{2.2cm} \texttt{Coreference Error} \\
    \\
    \hline
    \\
    \multicolumn{2}{c}{\parbox{16cm}{One of the biggest TV events of all time is being reimagined for new audiences. "Roots," the epic miniseries about an African-American slave and his descendants, had a staggering audience of over 100 million viewers back in 1977. Now A\&E networks are remaking the miniseries, to air in 2016. A\&E, Lifetime and History (formerly the History Channel) announced Thursday that the three networks would simulcast a remake of the saga [...]}} \\
    \\
    \vspace{0.2cm}
    A\&E, lifetime and history will simulcast a new "roots" in 2016. The original miniseries drew more than 100 million viewers in 1977. Levar Burton, who played Kunta Kinte in the original, will co-executive produce. & ``Roots,'' the epic miniseries about an african-american slave and his descendants , had a staggering audience of over 100 million viewers back in 1977. Now A\&E, lifetime and history (formerly the history channel) announced Thursday. Producers will consult scholars in african and african-american history for added authenticity.
    \\
    \hspace{3.5cm} \texttt{No Error} & \hspace{3.5cm} \texttt{Incoherent} \\
    \bottomrule
    \end{tabular}
    }
    \caption{\textbf{Two selected examples from the NND Summarization datasets.} For a given document, two candidates are provided: No Error (left) and Error (right). The top example is from the FRANK NND, and the bottom from the SummEval NND.}
    \label{appendix:nnd_ex_summ}
\end{table*}

\end{document}